\documentclass[10pt,twocolumn,letterpaper]{article}

\usepackage{multirow}
\usepackage[utf8]{inputenc} 
\usepackage[T1]{fontenc}    
\usepackage{booktabs}       
\usepackage{amsfonts}       
\usepackage{nicefrac}       
\usepackage{microtype}      
\usepackage{lipsum}
\usepackage{csquotes}
\usepackage{paralist,fancyhdr}
\usepackage[ruled, vlined, linesnumbered]{algorithm2e}
\usepackage{algorithmic}
\usepackage{amsthm}
\usepackage{mathrsfs} 
\usepackage{balance}
\usepackage{float}
\usepackage{stfloats}
\usepackage{flushend}
\usepackage{indentfirst}
\usepackage{subfigure}
\usepackage{colortbl}

\usepackage[pagenumbers]{iccv} 

\newcommand{\algoname}{DDE\xspace}

\newcommand\figref[1]{Fig.~\ref{#1}}
\newcommand\tabref[1]{Table~\ref{#1}}
\newcommand{\algref}[1]{Algorithm~\ref{#1}}

\newcommand\equref[1]{Eq.~\ref{#1}}
\newcommand\appref[1]{Suppl.~\ref{#1}}
\newcommand\secref[1]{Sec.~\ref{#1}}
\newcommand{\fakeparagraph}[1]{\noindent\textbf{#1}}

\definecolor{mygray}{rgb}{0.9, 0.9, 0.9}

\definecolor{iccvblue}{rgb}{0.21,0.49,0.74}
\usepackage[pagebackref,breaklinks,colorlinks]{hyperref}

\title{Preference Alignment for Diffusion Model via Explicit\\ Denoised Distribution Estimation}

\author{Dingyuan Shi,~Yong Wang\footnotemark[2]\;,~Hangyu Li,~Xiangxiang Chu \\
Alibaba Group \\
\small \texttt{dingyuan.shi@outlook.com, wangyong.lz@alibaba-inc.com} \\
\small \texttt{cyjdlhy@gmail.com, chuxiangxiang.cxx@alibaba-inc.com}
}

\begin{document}
\maketitle
{
\renewcommand{\thefootnote}{\fnsymbol{footnote}}
\footnotetext[2]{Corresponding Author}
}

\begin{abstract}
Diffusion models have shown remarkable success in text-to-image generation, making preference alignment for these models increasingly important. The preference labels are typically available only at the terminal of denoising trajectories, which poses challenges in optimizing the intermediate denoising steps. In this paper, we propose to conduct Denoised Distribution Estimation (DDE) that explicitly connects intermediate steps to the terminal denoised distribution. Therefore, preference labels can be used for the entire trajectory optimization. To this end, we design two estimation strategies for our DDE. The first is stepwise estimation, which utilizes the conditional denoised distribution to estimate the model denoised distribution. The second is single-shot estimation, which converts the model output into the terminal denoised distribution via DDIM modeling. Analytically and empirically, we reveal that DDE equipped with two estimation strategies naturally derives a novel credit assignment scheme that prioritizes optimizing the middle part of the denoising trajectory. Extensive experiments demonstrate that our approach achieves superior performance, both quantitatively and qualitatively. 
\end{abstract}
\section{Introduction}
\label{sec:intro}
Diffusion models have achieved remarkable success in text-to-image generation \cite{NeurIPS20Ho, ICLR21Songddim, CVPR22Rombach}. 
A key challenge in generative modeling is alignment \cite{CoRR24Liu, CoRR23Wang}, which focuses on improving a model's ability to better align with human preferences.
Alignment training has been extensively explored in the context of large language models \cite{CoRR20Stiennon, CoRR19Ziegler, NeurIPS22Ouyang}. 
Initially driven by Reinforcement Learning from Human Feedback (RLHF) \cite{ICLR24Lightman}, alignment techniques have evolved to many other approaches \cite{CoRR23RRHFYuan, ACL24Amini, NeurIPS24Wu} such as Direct Preference Optimization (DPO) \cite{NeurIPS23Rafailov, AISTATS24Azar, NeurIPS24Meng}. 
The latter has gained significant attraction, inspiring a series of subsequent studies \cite{ICML24Chen, CoRR23RRHFYuan, AISTATS24Azar}.

Despite the variety of emerging approaches, few studies have attempted to adapt DPO to text-to-image diffusion models. 
The primary challenge lies in the terminal-only issue of preference labels. 
That is, human evaluators can only provide preference labels for the final \textit{noiseless} output of generative models. 
In contrast, diffusion models generate images progressively, producing large numbers of \textit{noisy} intermediate results that are difficult to label. 
This raises the question: 
\textbf{How to optimize each \textit{intermediate} denoising step with \textit{terminal} preference labels only?}

Recent studies tend to solve this terminal-only issue from the perspective of credit assignment \cite{ICLR16Lillicrap, Arxiv23Mnih}. \ie, viewing the terminal preference signals as rewards and designing a scheme to distribute rewards among denoising steps.
These methods can be mainly divided into two categories.
One relies on auxiliary models, such as reward models \cite{CoRR23DPOK, ICLR24Black} or noisy evaluators \cite{CoRR24Liang}. 
These models essentially learn a weighting function to adaptively assign credit from the terminal reward to each denoising step. 
However, this approach introduces additional training complexity, undermining the simplicity of DPO.
The other requires hand-craft schemes, maintaining the simplicity of DPO but limiting the ability to perform effective credit assignments. 
These methods typically rely on simple strategies, such as uniform assignment \cite{CVPR24Yang, CVPR24Wallace} or discounted assignment \cite{ICML24Yang} (\ie placing more weight on the initial denoising steps), which may restrict the alignment potential of the model.

\begin{figure}[tbp]
\centering 
\setlength{\abovecaptionskip}{0.2cm}
\setlength{\belowcaptionskip}{-0.5cm}
\includegraphics[width=0.8\linewidth]{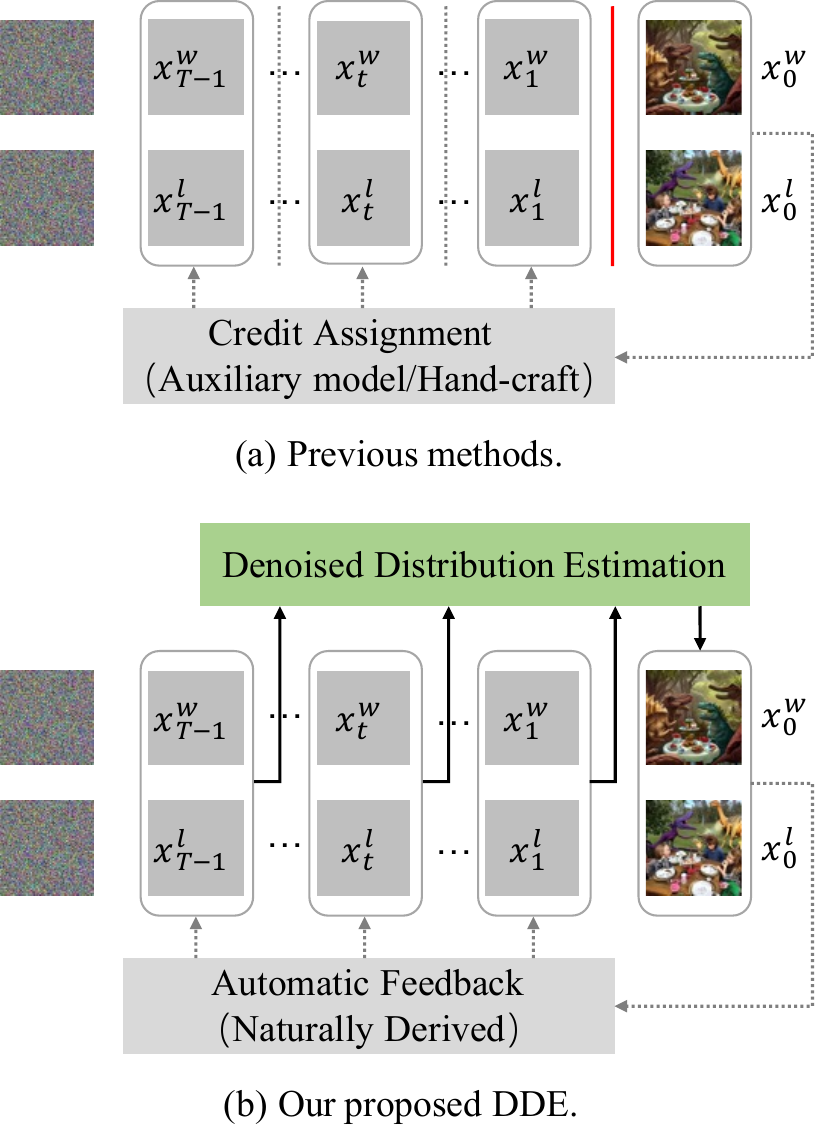}
\caption{
Comparison between previous methods and our DDE. 
The superscripts ``$w$'' and ``$l$'' denote winning and losing samples of a preference pair, respectively.
Previous methods ignore the connections among denoising steps, hence making the optimization heavily rely on credit assignment scheme of terminal preferences signals. 
In contrast, our DDE approach explicitly estimates the terminal distribution from any given step $t$, thereby naturally deriving a scheme that enables direct optimization for the preference labels.}
\label{fig:cmp}
\end{figure}

In this paper, we propose a novel approach termed \underline{D}enoised \underline{D}istribution \underline{E}stimation (DDE).
Instead of following the conventional perspective of credit assignment, we solve the problem from a novel angle, \ie, estimating the terminal denoised distribution explicitly (see \figref{fig:cmp}).
To achieve this, we design two strategies, namely \textit{stepwise} estimation and \textit{single-shot} estimation.
The stepwise strategy uses the ground-truth conditional distribution ($q(x_{t-1}|x_t, x_0)$) to estimate the model distribution ($p_\theta(x_{t-1}|x_t)$).
We also adopt a series of calibration coefficients for further accuracy.
This strategy helps directly estimate the distribution of $x_t$ before the sampled training step $t$.
The single-shot strategy utilizes DDIM modeling to directly estimate the terminal distribution ($p_\theta(\hat{x}_0|x_t)$) based on an intermediate noisy latent state.
This helps retain the model calculation pass only once, which can then be leveraged during training.

We analyze our \algoname, revealing that it naturally derives a credit assignment scheme that effectively prioritizes different denoising steps. 
Specifically, our estimation strategies introduce correction terms and step-specific coefficients, which lead to greater credit being assigned to the middle portion of the denoising trajectory. 
This scheme derivation is much easier than those relying on auxiliary models and much more powerful than simple hand-craft schemes.

We evaluate our method on both SD15 and SDXL quantitatively and qualitatively. 
The results demonstrate that compared with all other methods without auxiliary models, our \algoname achieves state-of-the-art performance. 
Specifically, our \algoname improves performance metrics of SD15 and SDXL by 3.3\%$\sim$6.7\% and 1.0\%$\sim$3.1\%, respectively.

Our contributions can be summarized as follows:
\begin{itemize}
\item 
We introduce \algoname, a novel DPO method tailored for diffusion models without any auxiliary model.
By directly etimating the terminal denoised distribution in terms of each step, \algoname naturally derives a novel automatic credit assignment scheme for terminal-only preference labels.

\item 
We design two novel estimation strategies namely stepwise estimation and single-shot estimation.
These strategies build connections from any denoising step $t$ to the terminal distribution, enabling preference optimization for all steps.
\item 
Experimental results show that \algoname achieves state-of-the-art performance both quantitatively and qualitatively, compared to existing methods without auxiliary models.
\end{itemize}

\begin{table}[htb]
\centering
\setlength{\belowcaptionskip}{-0.5cm}
\begin{tabular}{cc}
\toprule
Notation & Description \\
\hline
$x_0$ & Noiseless sample \\
$x_t, x_T$ & Noisy sample and pure noise \\
$\mu(x_t, x_0)$ & Expectation of $q(x_{t-1}|x_t, x_0)$ \\
$p_\theta(x_{t-1}|x_{t})$ & Estimated distribution of $x_{t-1}$ by DDPM \\
$\mu_\theta(x_t, t)$ & Expectation of $p_\theta(x_{t-1}|x_t)$ by DDPM \\
$p_\theta(\hat{x}_{t'}|x_t)$ & Estimated distribution of $x_{t'}$ by DDIM \\
$\hat{\mu}_{\theta, t'}(x_t, t)$ & Expectation of $p_\theta(\hat{x}_{t'}|x_t)$ by DDIM \\
\bottomrule
\end{tabular}
\caption{Notations}
\label{tab:notation}
\end{table}

\section{Preliminary}
\label{sec:pre}
\fakeparagraph{Diffusion Models.}
Traditional Denoising Diffusion Probabilistic Model (DDPM) \cite{NeurIPS20Ho} defines a forward process, which incrementally transforms a noiseless image or its latent \cite{CVPR23Blattmann, Alg23Scribano} (denoted as $x_0$) to pure noise (denoted as $x_T$) by injecting Gaussian noise as:
\begin{equation}
q(x_t|x_{t-1}) = \mathcal{N}(x_t; \sqrt{\alpha_t} x_{t-1}, \beta_t \mathbf{I}),
\label{equ:ddpm-forward-base}
\end{equation}
where $\beta_t$ and $\alpha_t\equiv 1 - \beta_t$ are hyperparameters of controlling the diffusion process.
The forward process can be expressed non-iteratively as:
\begin{equation}
q(x_t|x_0) = \mathcal{N}(x_t;\sqrt{\bar{\alpha}_t} x_0, (1 - \bar{\alpha}_t) \mathbf{I}),
\label{equ:ddpm-forward-fast}
\end{equation}
where $\bar{\alpha}_t = \prod_{i=1}^{t} \alpha_i$ denotes the cumulative product of the scaling factor $\alpha_t$.

DDPM requires a model to estimate the denoising distribution $p_\theta(x_{t-1}|x_t)$, which enables it to reverse the diffusion process and convert noise back into images \cite{ICLR21Songsm, JMLR05Aapo}. 
The training objective involves learning the distribution of $x_{t-1}$ conditioned on $x_0$, modeled as a Gaussian distribution:
\begin{equation}
q(x_{t-1}|x_t, x_0) = \mathcal{N}(x_{t-1}; \mu(x_t, x_0), \frac{1 - \bar{\alpha}_{t-1}}{1-\bar{\alpha}_t}\beta_t \mathbf{I}),
\label{equ:ddpm-backward}
\end{equation}
where $\mu(x_t,x_0) = \frac{\sqrt{\bar{\alpha}_{t-1}}\beta_t}{1 - \bar{\alpha}_t}x_0 + \frac{\sqrt{\alpha_t}(1 - \bar{\alpha}_{t-1})}{1 - \bar{\alpha}_t}x_t$. 
By minimizing the KL-divergence between model denoised distribution $p_\theta(x_{t-1}|x_t)$ and the true conditional denoised distribution above, the diffusion model learns to denoise the randomly sampled noise $x_T$ to $x_0$ progressively.

DDPM is constrained to denoise samples one step at a time.
In contrast, Denoising Diffusion Implicit Model (DDIM) \cite{ICLR21Songddim}  incorporates a subset of steps, redefining the denoising process as:
\vspace{-2mm}
\begin{equation}
p_\theta(\hat{x}_{t'}|x_t) = \mathcal{N}(\hat{x}_{t'}; \hat{\mu}_{\theta, t'}(x_t, t), \sigma_t \mathbf{I}),
\label{equ:ddim-backward}
\end{equation}
where $t'$ and $t$ are two consecutive steps within the subset of the original DDPM denoising steps, $\hat{\mu}_{\theta, t'}(x_t, t)=\sqrt{\bar{\alpha}_{t'}}(\frac{x_t - \sqrt{1 - \bar{\alpha}_t}\epsilon_\theta(x_t, t)}{\sqrt{\bar{\alpha}_t}}) + \sqrt{1 - \bar{\alpha}_{t'} - \sigma_t^2} \epsilon_\theta(x_t, t)$ and $\sigma_t = \sqrt{\frac{1 - \bar{\alpha}_{t'}}{1 - \bar{\alpha}_t} \beta_t}$.

\fakeparagraph{Direct Preference Optimization.}
DPO \cite{NeurIPS23Rafailov} aims to optimize generative models to better align with human preferences.
This objective is achieved by establishing a connection between the reward model and the generative model, resulting in the following loss function:
\begin{equation}
    \mathcal{L}_{\text{DPO}} = \mathbb{E}_{x^w, x^l\sim \mathcal{D}}[-\log \sigma(\beta(\log \frac{p_\theta(x^w)}{p_{ref}(x^w)} -  \log \frac{p_\theta(x^l)}{p_{ref}(x^l)}))]
\label{equ:dpo-naive}
\end{equation}
where $p_\theta$ and $p_{ref}$ represent the target and reference model respectively.
The samples $x^w$ and $x^l$ constitute a preference pair, where $x^w \succ x^l$ indicates that the winning sample $x^w$ is preferred over the losing sample $x^l$.
The sigmoid function is denoted as $\sigma$.  
In this loss formulation, the differential term ($\log \frac{p_\theta(x^w)}{p_{ref}(x^w)} - \log \frac{p_\theta(x^l)}{p_{ref}(x^l)}$) serves to adjust the model to increase the probability of generating the winning sample ($x^w$), while decreasing that of the losing sample  ($x^l$).
The fractional components $\frac{p_\theta}{p_{ref}}$ keep the optimization close to the reference model, thus preventing excessive deviations.
\tabref{tab:notation} summarizes frequently used notations of the paper.

\begin{figure*}[t]
\centering
\setlength{\abovecaptionskip}{0.2cm}
\setlength{\belowcaptionskip}{-0.5cm}
\includegraphics[width=1.0\linewidth]{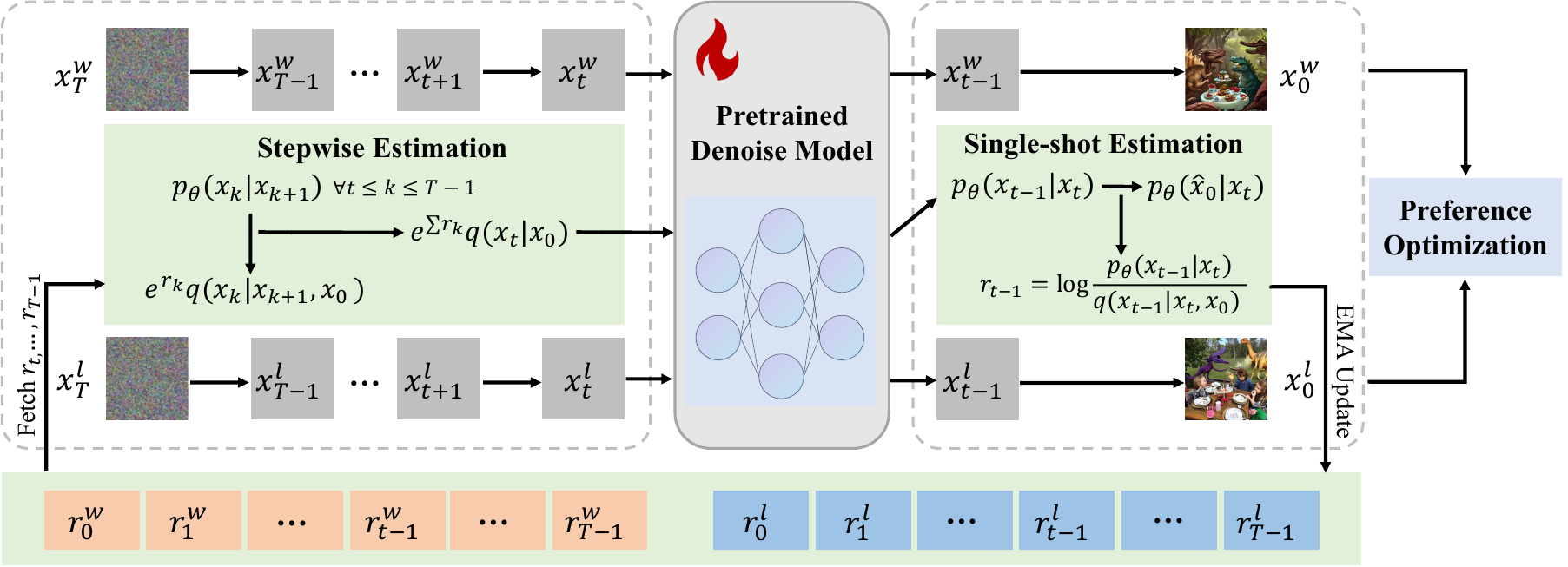}
\caption{
Overall framework of \algoname. 
The training process is outlined as follows: 1) Sample random noises  ($x_T^w, x_T^l$), winning and losing samples ($x_0^w, x_0^l$) and a denoising step $t$;
2) Conduct stepwise estimation from $T\rightarrow t$. By using $\exp\{r_k\} q(x_{k}|x_{k+1}, x_0)$ as an estimation to $p_\theta(x_k|x_{k+1})$ for all $t\le k \le T-1$, the cumulative product of denoising steps from $T$ to $t$ is estimated as $\exp\{\sum_{k=t}^{T-1} r_k\} q(x_t|x_0)$;
3) Apply single-shot estimation from $t \rightarrow 0$. By using DDIM, $p_\theta(x_{t-1}|x_t)$ is converted to $p_\theta(\hat{x}_{0}|x_t)$ with one single model calculation;
4) Leverage the preference label on $x_0$ for training.
Additionally, in step 3, $p_\theta(x_{t-1}|x_t)$ is used to calculate non-gradient calibration coefficients $r_{t-1} = \log \frac{p_\theta(x_{t-1}|x_t)}{q(x_{t-1}|x_t, x_0)}$. These coefficients are updated using an exponential moving average for subsequent iterations.
}
\label{fig:framework}
\end{figure*}

\section{Method}
\label{sec:method}
We present our Denoised Distribution Estimation (\algoname) approach.
The framework is shown in ~\figref{fig:framework}. 
Distinct from previous studies, which try to solve the terminal-only issue of preference label from the perspective of credit assignment, our \algoname explores a novel approach.
It directly estimates the terminal denoised distribution $p_\theta(x_0)$, from any intermediate  step $p_\theta(x_{t-1}|x_t)$. 
This allows for the optimization across all steps by using terminal preference labels only. 
To achieve this, we can start from representing terminal $p_\theta(x_0)$ in terms of all steps $p_\theta(x_t)$ as follows:
\vspace{-2mm}
\begin{equation}
\begin{aligned}
p_\theta(x_0) &= \int_{x_{1:T}} p_\theta(x_{0:T}) dx_{1:T} \\
&= \int_{x_{1:T}} q(x_T)p_\theta(x_{T-1}|x_T)...p_\theta(x_0|x_1) dx_{1:T} \\
\end{aligned}
\label{equ:framework}
\end{equation}
where $q(x_T)$ is a Gaussian distribution and $p_\theta(x_{t-1}|x_t)$ denotes the learned distribution which constitutes the denoising trajectory.
Subsequently, we discuss the necessity and practice of our \algoname.

\fakeparagraph{Why use Denoised Distribution Estimation (\algoname)?}
The most straightforward approach would involve calculating the entire denoising trajectory in \equref{equ:framework},  which would allow for the direct use of terminal preference labels for optimization.
However, each $p_\theta$ term in \equref{equ:framework} indicates a model calculation pass.
This straightforward approach would incur prohibitive training costs due to the iterative nature of the generation process. 
The training feasibility requires a constrained number of model calculations.
Therefore, the final denoised distribution must be estimated with a reduced number of model passes, ideally within a single one.
This motivates us to build estimations for the entire denoising trajectory.

\fakeparagraph{\algoname consists of two estimation strategies.}
Let $t$ denote a sampled denoising step to be optimized during training, and it is required to calculate $p_\theta(x_{t-1}|x_t)$ for optimization.
Consequently, the denoising trajectory can be naturally split into two segments: from $T$ to $t$ and from $t$ to $0$, denoted as $T\rightarrow t$ and $t \rightarrow 0$, respectively.
It is important to note that these two segments are different.
The segment $T\rightarrow t$ is prior to the model pass (which happens at sampled $t$), hence it requires for an estimation without any $p_\theta$ (\ie model calculation).
The segment $t\rightarrow 0$ is posterior to model calculation, so we need to estimate the terminal distribution $p_\theta(\hat{x}_0|x_t)$ with limited or preferably only one single model pass.
We propose two distinct strategies for each segment:
\begin{itemize}
    \item \textit{Stepwise estimation for segment $T \rightarrow t$.} 
    We estimate each step term $p_\theta(x_t|x_{t+1})$ by $q(x_t|x_{t+1}, x_0)$ (defined as \equref{equ:ddpm-backward}).
    To further ehance the estimation accuracy, we incorporate a series of coefficients for calibration.
    \item \textit{Single-shot estimation for segment $t\rightarrow 0$.} We directly estimate the terminal denoised distribution of $x_0$ from that of $x_t$ by using DDIM with a single model pass.
\end{itemize}

\subsection{Stepwise Estimation for Segment $T\rightarrow t$}
The stepwise estimation uses $q(x_k|x_{k+1}, x_0)$ (defined as \equref{equ:ddpm-backward}) to estimate $p_\theta(x_k|x_{k+1})$ for all $k$ from $T$ to $t$.
This approach is justifiable as the model is optimized during pretraining to minimize the KL-divergence between the above two distributions \cite{NeurIPS20Ho}.
To further enhance the accuracy of the estimation, we adopt a series of constant calibration coefficients $r_k$.
We will introduce its calculation in subsequent paragraph.
If multiply $q(x_k|x_{k+1}, x_0)$ with an appropriate $\exp\{r_k\}$, it will become closer to $p_\theta(x_k|x_{k+1})$ for all $k\in \{t, ..., T-1\}$.
Therefore, the stepwise estimation substituting all $p_\theta(x_k|x_{k+1})$ with $\exp \{r_k\} q(x_k|x_{k+1}, x_0)$ yields (the comprehensive derivation is provided in \appref{subapp:esa}):
\vspace{-2mm}
\begin{equation}
p_\theta(x_0) = \exp\{\sum_{k=t}^{T-1} r_k\} \int_{x_{1:t}} q(x_t|x_0) \prod_{k=t}^1 p_\theta(x_{k-1}|x_k) dx_{1:t}
\label{equ:esa}
\end{equation}
By adopting the stepwise estimation, the denoising segment from $T$ to $t$ can be simplified to a single term $q(x_t|x_0)$, with an additional correction term $\exp\{\sum_{k=t}^{T-1} r_k\}$. Term $q(x_t|x_0)$ can be computed using \equref{equ:ddpm-forward-fast}.
\textit{In other words, the stepwise estimation exempts any extra model pass within segment $T\rightarrow t$.}
The correction term can be calculated based on a series of calibration coefficients, which we will introduce its calculation next.

\fakeparagraph{Calculation of calibration coefficients $r_k$.}
Since we want $\exp\{r_k\}q(x_k|x_{k+1}, x_0) \approx p_\theta(x_k|x_{k+1})$, the $r_k$ should equals to $\log \frac{p_\theta(x_k|x_{k+1})}{q(x_k|x_{k+1}, x_0)}$.
However, $r_k$ is primarily attributed to $p_\theta(x_k|x_{k+1})$ for $k \in \{t,...,T-1\}$, which remains inaccessible when training $p_\theta(x_{t-1}|x_t)$.
To avoid extra model inferences when obtaining the set of $p_\theta(x_k|x_{k+1})$, we use a series of non-gradient coefficients for calibration.
Specifically, we maintain an array of length $T$ for recording $r_k$ and employ exponential moving average (EMA) to update this value throughout the training process.
Empirically, we observe that these coefficients converge quickly (as shown in \figref{subfig:ct_training}), and this EMA manner does not adversely affect the training process.

Subsequently, we introduce our single-shot estimation which elaborates the term $\prod_{k=t}^1 p_\theta(x_{k-1}|x_k)$ in \equref{equ:esa}. 

\subsection{Single-shot Estimation for Segment $t \rightarrow 0$}
As we have mentioned above, the segment from $t$ to $0$ requires for estimating terminal distribution $p_\theta(\hat{x}_0|x_t)$ with limited or single model pass. 
Considering the power of DDIM on calculation number reduction, we try to adopt DDIM modeling for single-shot estimation. 
We first conduct integral of term $\int_{x_{1:t-1}}\prod_{k=t}^1 p_\theta(x_{k-1}|x_k) dx_{1:t-1}$ in \equref{equ:esa} and get $p_\theta(x_0|x_t)$.
Therefore, \equref{equ:esa} can be rewritten as follows:
\vspace{-2mm}
\begin{equation}
\begin{aligned}
p_\theta(x_0) & \approx \exp\{\sum_{k=t}^{T-1} r_k\} \int_{x_t} q(x_t|x_0) p_\theta(x_0|x_t) dx_t  \\
& = \exp\{\sum_{k=t}^{T-1} r_k\} 
 \mathbb{E}_{x_t \sim q(x_t|x_0)}[p_\theta(x_0|x_t)].
\end{aligned}
\label{equ:all}
\end{equation}
By viewing $x_t$ and $x_0$ as consecutive steps in a DDIM subset of the original denoising steps, we have:
\begin{equation}
\begin{aligned}
\hat{\mu}_{\theta, t'=0}(x_t, t)= & \sqrt{\bar{\alpha}_0}(\frac{x_t - \sqrt{1 - \bar{\alpha}_t}\epsilon_\theta(x_t, t)}{\sqrt{\bar{\alpha}_t}}) \\ 
& + \sqrt{1 - \bar{\alpha}_0 - \sigma_t^2} \epsilon_\theta(x_t, t)
\end{aligned}
\label{equ:single-shot}
\end{equation}
where $\sigma_t = \sqrt{\frac{1 - \bar{\alpha}_{t-1}}{1 -\bar{\alpha}_t} \beta_t}$.
Therefore, $p_\theta(\hat{x}_0|x_t) = \mathcal{N}(\hat{x}_0; \hat{\mu}_{\theta, t'=0}(x_t, t), \sigma_t \mathbf{I})$, and the $p_\theta(x_0|x_t)$ in \equref{equ:all} can be estimated with only one single pass of the model.

\fakeparagraph{Why use one single calculation?}
We aim to enhance training efficiency by utilizing a minimal number of model calculations when performing single-shot estimation from $t$ to $0$.
The aforementioned strategy employs only one calculation.
Concerns may arise about the sufficiency of a single calculation, particularly when $t$ is large, as using a single calculation to reach $0$ through DDIM typically does not produce clear images.
However, we argue that one calculation step is adequate.
Our focus is on preference optimization, not text-to-image training.
The relative differences between preference pairs are more important than the absolute differences between $p_\theta(x_0|x_t)$ and $p_\theta(\hat{x}_0|x_t)$.
While one single calculation might lack accuracy, the relative differences are sufficient for effective preference optimization.
Experimental results indicate that the performance is adequate, and there is no necessity to increase the number of calculations during single-shot estimation.

\subsection{Combining Two Estimation Strategies}
Building on the above two estimation strategies, the loss can be derived as follows:
\vspace{-2mm}
\begin{equation}
\begin{aligned}
& \mathcal{L}_{\text{\algoname}}  \\
&= \mathbb{E}_{x^w_0, x^l_0, x^w_t \sim q(x^w_t|x^w_0), x^l_t \sim q(x^l_t|x^l_0)}[-\log \sigma (\beta(  \\
& -||x_0^w - \hat{\mu}_{\theta, t'=0}(x_t^w, t)||_2^2 + ||x_0^w - \hat{\mu}_{ref, t'=0}(x_t^w, t)||_2^2  \\
& + ||x_0^l- \hat{\mu}_{\theta, t'=0}(x_t^l, t)||_2^2 - ||x_0^l - \hat{\mu}_{ref, t'=0}(x_t^l, t)||_2^2  \\
& +\sum_{k=t}^{T-1} (r_{\theta,k}^w - r_{ref, k}^w - r_{\theta, k}^l + r_{ref, k}^l)
))],
\end{aligned}
\label{equ:dpo-loss-approx}
\end{equation}
where $\hat{\mu}_\theta$ and $\hat{\mu}_{ref}$ denote using target and reference model to estimate $x_0$ by $x_t$ via DDIM respectively, as introduced in \secref{sec:pre}.
The detailed derivation can be found at  \appref{subapp:loss}.

We provide an intuitive explanation for each term involved in our formulation.
For Mean Square Error (MSE) terms related to $\theta$, $||x_0^w - \hat{\mu}_{\theta, t'=0}(x_t^w, t)||_2^2$ guides the optimizer to increase the output probability of $p_\theta(x_0^w|x^w_t)$, while $||x_0^l- \hat{\mu}_{\theta, t'=0}(x_t^l, t)||_2^2$ decreases the model output samples like $x_0^l$.
MSE terms with respect to the reference model penalize deviations from the reference, with large deviations pushing the optimizer toward the saturation region of the negative log-sigmoid function, thus weakening the optimization. 
The terms containing coefficients $r_k$ correct these deviations, modifying the MSE terms in the negative log-sigmoid function, which adjusts the optimization weight. Note that although $r_k$ is a series of non-gradient constants, they can still affect the training loss and parameter optimization, as will be discussed in \secref{subsec:dde-prioritize}.

The whole training process is detailed in \algref{alg:dde-training}.
For simplicity, we use $w/l$ in the superscripts denoting calculation on both winning and losing samples, respectively.
Lines 5 and 6 illustrate our stepwise and single-shot estimation strategies, respectively.
Lines 9 and 10 represent our non-gradient value updating of our calibration coefficients which support the stepwise estimation.

\subsection{Discussion}
\label{subsec:dde-prioritize}

\fakeparagraph{\algoname prioritizes optimizing middle part of denoising steps.}
To further clarify the novelty of our method, we also provide analysis from the perspective of credit assignment, following the convention of existing work.
We find our stepwise and single-shot estimations weaken the optimization of steps around $0$ and $T$, respectively.

The stepwise estimation strategy weakens the optimization near $0$.
This is because it introduces correction terms to the loss function, which becomes larger as denoising step approaching to $0$ (see \figref{subfig:correction-coeff} in \secref{subsec:correction}).
A larger correction term pushes the MSE terms closer into the gradient saturation region of the negative log-sigmoid function, thus reducing the optimization effectiveness. In contrast, the single-shot estimation weakens the optimization near $T$.
Single-shot estimation introduces weight coefficients from DDIM modeling with maximal values near $T$, which magnify even a slight difference between the training and the reference model.
It will push the term into the gradient saturation region of loss function resulting in attenuating the optimization. In particular, two strategies weaken both sides of the denoising trajectories, hence it naturally derives a credit assignment scheme prioritizing the optimization of the middle steps of denoising trajectories.
Please refer to \secref{subsec:correction} for more discussion.

\fakeparagraph{\algoname derives a credit assignment scheme naturally.}
Unlike previous methods that construct the scheme from the perspective of credit assignment and modeling the denoising process as a Markov sequential decision problem, our \algoname explicitly estimates the learned denoised distribution.
This provides deeper insight into the denoising trajectory and naturally derives a finer credit assignment scheme.
This methodology is exempted from the extra workload of auxiliary model training, while also avoiding the suffering of the rough granularity of hand-craft methods.

\vspace{-2mm}
\begin{algorithm}[h]
\SetAlgoLined
\caption{Denoised Distribution Estimation}
\label{alg:dde-training}
\newcommand{\nonl}{\renewcommand{\nl}{\let\nl\oldnl}}
\KwIn{Target pretrained model $\theta$, reference model $ref$, preference dataset $\mathscr{D}$, initialized coefficient array $r_\theta^{w/l}[0,...,T-1]\leftarrow 0$, $r_{ref}^{w/l}[0,...,T-1]\leftarrow 0$}
\KwOut{Finetuned model}
\For{$\text{training\_step}$=1,2,...,MAX\_STEP}{
    $(x^w_0, x^l_0) \sim \mathscr{D}$, $t \sim Uniform\{1,...,T\}$ \\
    $(x_T^w, x_T^l) \sim Gaussian(\mathbf{0},\mathbf{I})$\\
    $(x^w_t, x^l_t) \leftarrow$ Add noise by \equref{equ:ddpm-forward-fast} \\
    $\triangleright$ Stepwise estimation \\
    {\nonl $\textit{correction\_term} \leftarrow \sum_{k=t}^{T-1} (r_{\theta}^w[k] - r_{ref}^w[k] - r_{\theta}^l[k] + r_{ref}^l[k])$}  \\
    $\triangleright$ Single-shot estimation \\
    {\nonl Calculate $\hat{\mu}_{\theta, t'=0}(x^{w/l}_t, t), \hat{\mu}_{ref, t'=0}(x^{w/l}_t, t)$ by \equref{equ:single-shot}} \\
    $\mathcal{L} \leftarrow$ Calculate loss by \equref{equ:dpo-loss-approx} with $\textit{correction\_term}$, $\hat{\mu}_{\theta, t'=0}(x^{w/l}_t, t)$ and $\hat{\mu}_{ref, t'=0}(x^{w/l}_t, t)$\\
    Apply back propagation and update based on $\mathcal{L}$ \\
    $r_\theta^{w/l}[t - 1] \leftarrow$ EMA-Update($\log \frac{p_\theta(x^{w/l}_{t-1}|x^{w/l}_t)}{q(x^{w/l}_{t-1}|x^{w/l}_t, x_0)}$) \\
    $r_{ref}^{w/l}[t - 1] \leftarrow$ EMA-Update($\log \frac{p_{ref}(x^{w/l}_{t-1}|x^{w/l}_t)}{q(x^{w/l}_{t-1}|x^{w/l}_t, x_0)}$) \\
}
\end{algorithm}

\section{Experiments}
\label{ref:exp}
\subsection{Implementation Detail}
We finetune the models using the Pick-a-Pic-V2 dataset \cite{dataset} with the popular text-to-image models, Stable Diffusion 1.5 (SD15) and Stable Diffusion XL (SDXL).
All parameters of the U-Net are trained.
SD15 and SDXL are trained for 2000 and 1500 iterations, respectively.
Training is conducted on 8 Nvidia H20 GPUs, each with 96GB memory, and a batch size of 2048. Please refer to \appref{subapp:detail} for more  details.

\begin{table}[htb]
\centering
\setlength{\belowcaptionskip}{-0.2cm}
\scalebox{0.82}{
\begin{tabular}{lccc}
\toprule
\textbf{Methods} & \textbf{CLIP} ($\uparrow$) &  \textbf{HPS} ($\uparrow$)  & \textbf{PS} ($\uparrow$)  \\
\midrule
SD15~\cite{CVPR22Rombach}        & 3.200 $\pm$ 0.657 & 2.622 $\pm$ 0.208 & 2.049 $\pm$ 0.131 \\
SFT          & 3.306 $\pm$ 0.685 & 2.714 $\pm$ 0.245 & 2.101 $\pm$ 0.137 \\
Uni~\cite{CVPR24Wallace, CVPR24Yang}         & 3.313 $\pm$ 0.748 & 2.703 $\pm$ 0.234 & 2.105 $\pm$ 0.134 \\
Disc~\cite{ICML24Yang}         & 3.317 $\pm$ 0.730 & 2.720 $\pm$ 0.248 & 2.083 $\pm$ 0.145 \\
\midrule
\rowcolor{mygray}\algoname & \textbf{3.414} $\pm$ \textbf{0.627} & \textbf{2.725} $\pm$ \textbf{0.216} & \underline{2.112 $\pm$ 0.137} \\
\rowcolor{mygray}\algoname-Single  & \underline{3.386 $\pm$ 0.722} & \underline{2.723 $\pm$ 0.204} & \textbf{2.117} $\pm$ \textbf{0.129} \\
\rowcolor{mygray}\algoname-Step  & 3.168 $\pm$ 0.671 & 2.600 $\pm$ 0.671 & 2.056 $\pm$ 0.123 \\
\bottomrule
\end{tabular}
}
\caption{Experimental results on SD15.
Our \algoname outperforms all baseline methods across all metrics.
DDE-Single employs single-shot estimation from $t$ to $0$ and stepwise estimation without correction terms from $T$ to $t$.
DDE-Step adopts stepwise estimation from $T$ to $t$ and performs preference optimization on $x_t$.
Employing only one of the estimation method results in a decrease in performance.}
\label{tab:validation}
\end{table}

\begin{figure*}[t]
\centering
\setlength{\abovecaptionskip}{0.2cm}
\setlength{\belowcaptionskip}{-0.5cm}
\includegraphics[width=1.0\linewidth]{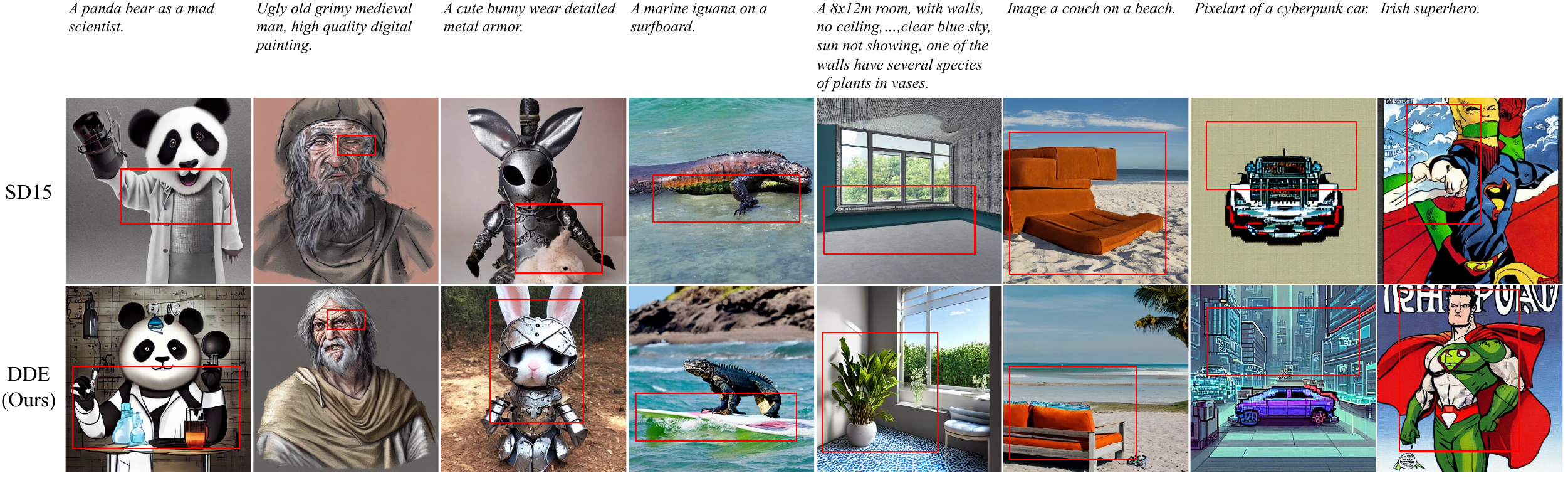}
\caption{Our model generates images with better detail, structure, and text-alignment, compared to the SD15 model.
Specifically, it can generate a panda bear with tubes and flasks, which better aligns with ``scientist'' in the prompt. 
Additionally, the head portrait has more accurate eye detail and the rabbit is \textit{wearing} an armor.  
We can generate images containing items as prompt requested (\eg surfboard, vases).
The structure can be maintained (\eg sofa and superhero) and the background is more detailed (\eg the cyberpunk cars).
}

\label{fig:cmp-sd15}
\end{figure*}

\subsection{Quantitative Validation}

To validate our approach quantitatively, we leverage three distinct models as preference annotators, including CLIP \cite{ICML21Radford}, HPS \cite{CoRR23Wu}, and PS \cite{NeurIPS23Kirstain}. Each model specializes in capturing different dimensions of human preference evaluation.
Based on these annotators, we train our proposed method (\algoname) in conjunction with several baseline methods, including Supervised Finetune (SFT), Uniform (Uni, \cite{CVPR24Wallace, CVPR24Yang}), and Discounted (Disc, \cite{ICML24Yang}).

\fakeparagraph{SD15 Models.}
The quantitative evaluation results are presented in \tabref{tab:validation}.
Our proposed approach achieves significant performance enhancements over the base models and surpasses all other baseline methods.
Specifically, our \algoname demonstrates consistent improvements of 6.7\% (CLIP), 3.9\% (HPS), and 3.3\% (PS) over SD15, respectively. 
These empirical results clearly demonstrate the effectiveness of our proposed method.

\fakeparagraph{SDXL Models.}
As shown in \tabref{tab:validation-sdxl}, our proposed algorithm, \algoname, demonstrates superior performance across all evaluated metrics. Specifically, \algoname enhances the performance of SDXL by 1.4\%, 1.0\%, and 3.1\%  across the respective evaluation metrics. These significant improvements underline the efficacy of our algorithm in achieving more accurate results compared to the existing standard baselines represented by SDXL. The consistent performance gains across diverse scenarios indicate the robustness and potential of \algoname for broader applications.

\begin{table}[h]
\setlength{\abovecaptionskip}{0.2cm}
\setlength{\belowcaptionskip}{-0.3cm}
\centering
\scalebox{0.87}{
\begin{tabular}{lccc}
\toprule
\textbf{Methods} & \textbf{CLIP} ($\uparrow$) & \textbf{HPS} ($\uparrow$)  & \textbf{PS} ($\uparrow$) \\
\midrule
SDXL~\cite{Arxiv23Podell}    & 3.664 $\pm$ 0.579 & 2.803 $\pm$ 0.172 & 2.154 $\pm$ 0.145 \\
Uni~\cite{CVPR24Wallace, CVPR24Yang}    & 3.699 $\pm$ 0.542 & 2.806 $\pm$ 0.168 & 2.191 $\pm$ 0.134 \\
Disc~\cite{ICML24Yang}  & 3.677 $\pm$ 0.540 & 2.742 $\pm$ 0.155 & 2.136 $\pm$ 0.130 \\
\midrule
\rowcolor{mygray}\algoname  & \textbf{3.715} $\pm$ \textbf{0.521} & \textbf{2.831} $\pm$ \textbf{0.175} & \textbf{2.224} $\pm$ \textbf{0.137} \\
\bottomrule
\end{tabular}}
\caption{Experimental results on SDXL demonstrate that \algoname consistently surpasses all baseline models across all evaluated metrics.}
\label{tab:validation-sdxl}
\end{table}

\fakeparagraph{Beat Ratio Comparison.}
Beyond evaluating the mean performance across the entire set of validation images, we conducted a comprehensive examination of the beat ratios to gain a more nuanced understanding of algorithm's effectiveness. As illustrated in \tabref{tab:beat-ratio}, our \algoname shows superior performance compared to SD15. Specifically, our method achieves beat ratios of $65.6\%$, $76.7\%$, $72.7\%$ in terms of CLIP, HPS, and PS, respectively. These results substantiate the robustness of \algoname, consistently outperforming other baseline methods across different evaluation scenarios.

\begin{table}[h]
\centering
\setlength{\abovecaptionskip}{0.2cm}
\setlength{\belowcaptionskip}{-0.5cm}
\scalebox{0.87}{
\begin{tabular}{lccc}
\toprule
\textbf{Methods} & \textbf{CLIP} ($\uparrow$) & \textbf{HPS} ($\uparrow$)  & \textbf{PS} ($\uparrow$) \\
\midrule
SFT & 0.5798 & 0.7362 & 0.6840 \\
Uni~\cite{CVPR24Wallace, CVPR24Yang} & 0.6043 & 0.6840 & 0.6656 \\
Disc~\cite{ICML24Yang} & 0.5767 & 0.7301 & 0.5951 \\
\midrule
\rowcolor{mygray}\algoname & \textbf{0.6564} & \textbf{0.7669} & \textbf{0.7270} \\
\bottomrule
\end{tabular}}
\caption{The comparison of beat ratios indicates that our DDE achieves the highest beat ratios across all validation metrics compared to other baselines.}
\label{tab:beat-ratio}
\end{table}

\begin{figure*}[t]
\setlength{\abovecaptionskip}{0.2cm}
\setlength{\belowcaptionskip}{-0.5cm}
\centering
\includegraphics[width=1.0\linewidth]{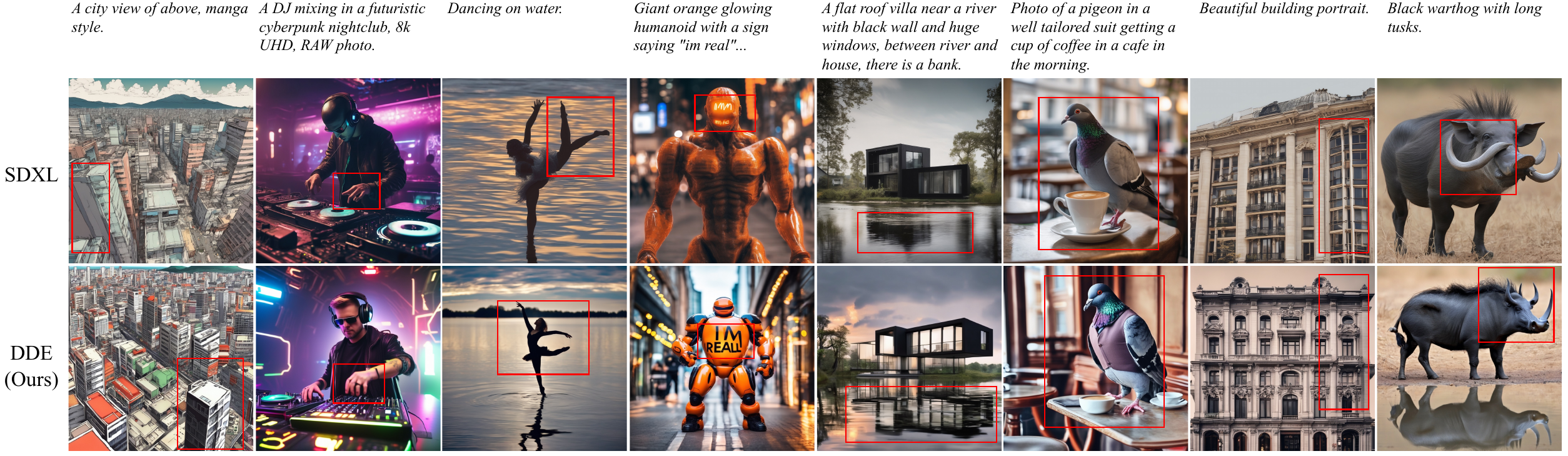}
\caption{Our model generates images with better detail, structure, text-alignment than SDXL model.
We can generate building retaining window details and human hands with the right number of fingers.
The dancing body structure can be kept and the requested text as well as the reflection in the water can be correctly generated.
We generate a pigeon \textit{wearing} a suit specified by the prompt.
}
\label{fig:cmp-sdxl}
\end{figure*}

\subsection{Qualitative Validation}
We also conduct a qualitative comparison between our DDE and the baseline models and provide some illustrative examples. More generated cases are available in \appref{subapp:quality}.

\fakeparagraph{SD15 Models.}
As shown in \figref{fig:cmp-sd15}, our model surpasses SD15 in generating images with superior details, structure, and text alignment.
For instance, the ``panda scientist'' produced by using our approach holds on flasks and tubes, exhibiting enhanced detail generation capability compared to the base models.
The head portrait generated by the base model suffers eye structure collapse, which our model successfully avoids. 
Furthermore, our model produces rabbit \textit{wearing} an armor that is more aligned with the given prompt.
Additionally, the generated image has a better view of the surfboard as the prompt request, and the room corner we produce displays more details and a clearer view.
The sofa and the superhero maintain more reasonable structures.
The cyberpunk car image generated by our \algoname shows a finer background than other baseline models.

\fakeparagraph{SDXL Models.}
Our model also enhances the SDXL model, as illustrated in \figref{fig:cmp-sdxl}.
For the city view generation, we maintain the detail of windows on the buildings more effectively than the base model.
Our approach accurately generates a human hand with the right number of fingers and a dancer with the correct body structure.
The text generated by our model is clearer, aligning precisely with the prompt. 
Moreover, the inverted reflections in water generated by our model exhibit more accurate structural details.
Our method is also more capable of following text prompts, with the pigeon \text{wearing} a suit as requested.

\begin{figure*}[h]
\centering
\setlength{\abovecaptionskip}{0.1cm}
\setlength{\belowcaptionskip}{-0.5cm}
\subfigure[]{
\includegraphics[width=0.33\linewidth]{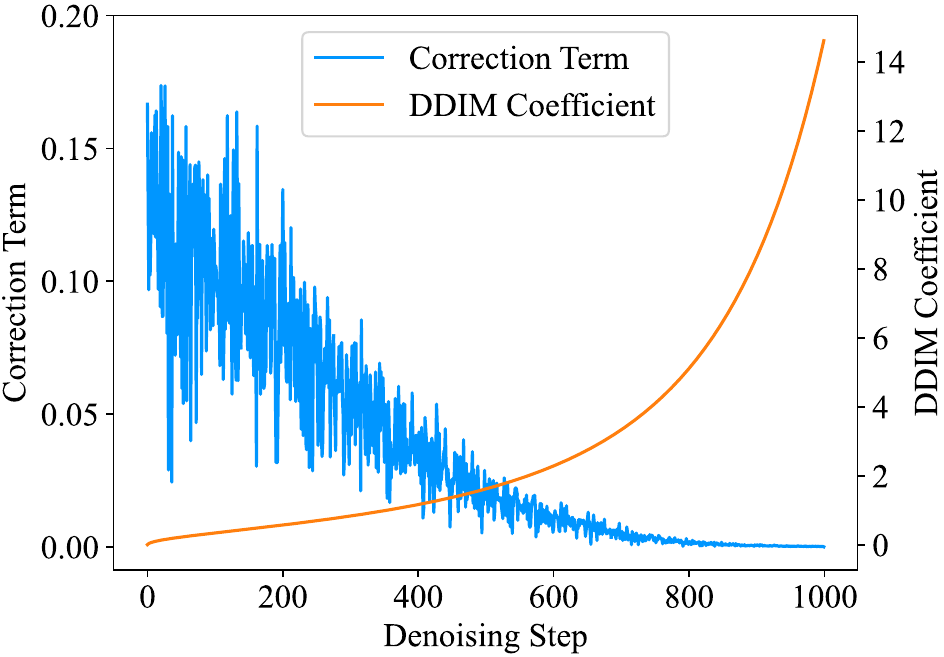}
\label{subfig:correction-coeff}
}
\subfigure[]{
\includegraphics[width=0.29\linewidth]{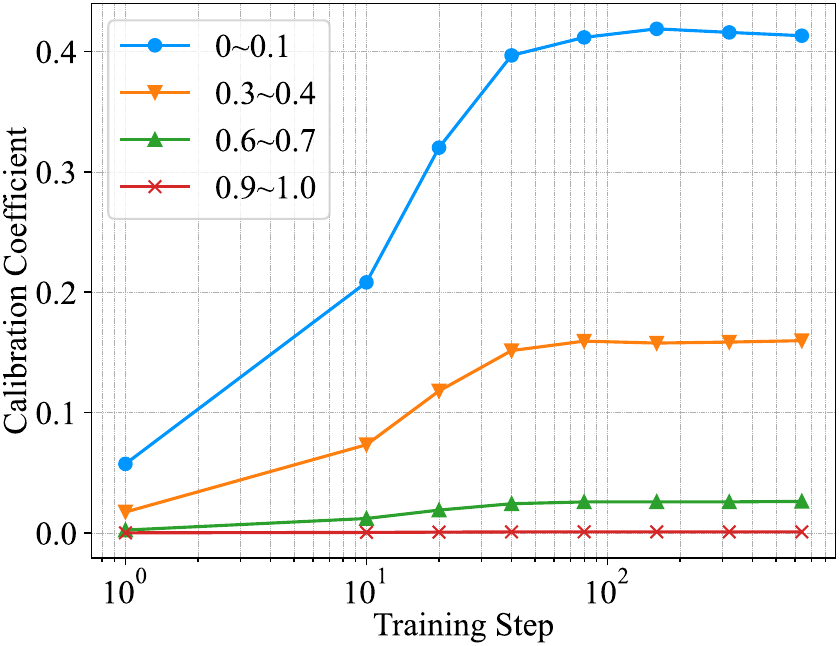}
\label{subfig:ct_training}
}
\subfigure[]{
\includegraphics[width=0.31\linewidth]{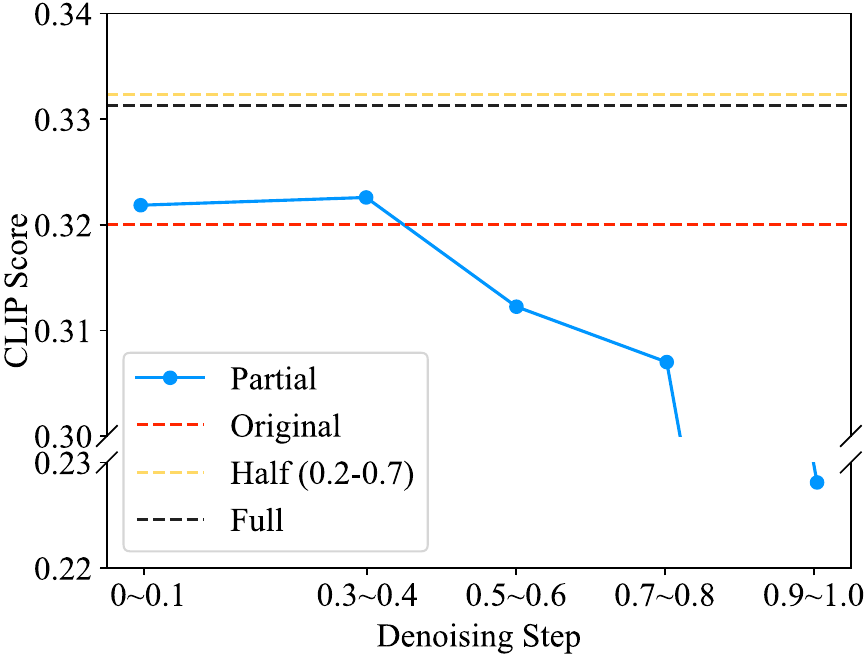}
\label{subfig:analysis}
}
\caption{(a) The values of the correction terms and DDIM coefficients, show that larger values result in weakening the effectiveness of denoising optimization at both boundaries. (b) The convergence behavior of the EMA update for calibration coefficients $r_t$, indicates stabilization within 100 training iterations. (c) A comparative analysis of optimization across various steps, reveals that optimizing only half of the steps yields performance comparable to that achieved by optimizing full steps.}
\end{figure*}

\subsection{Ablation study}
\label{subsec:ablation}
We conduct the ablation study using following methods.
\begin{itemize}
\item 
\algoname-Step: stepwise estimation from $T$ to $t$, then conduct preference optimization directly on $x_t$.
\item 
\algoname-Single: stepwise estimation from $T$ to $t$ without coefficients $r_t$, and single-shot estimation from $t$ to $0$.
\end{itemize}
It is important to note that single-shot estimation cannot be applied prior to $t$, as this would result in substantial computational expenses due to multiple model passes.

The results are shown in the bottom section of \tabref{tab:validation}.
The performance decline validates the effectiveness of both estimation strategies.
Furthermore, the DDE-Step exhibits suboptimal performance, suggesting that the direct application of terminal preference labels at intermediate step $t$ does not improve or even damage alignment training.
This can be attributed to the fact that the preference labels on the terminal noiseless sample do not mean the whole denoising trajectory is preferred.
The performance drop of DDE-Single validates the effectiveness of our coefficients $r_t$.
Removing the calibration coefficients not only changes the denoising trajectory distribution prior to $t$, but also modifies the credit assignment scheme, thus impairs the performance.

\subsection{More Analysis}
\label{subsec:exp-prioritize}

In this part, we provide some deeper insight into how our method prioritizes the middle step and why it works better than other schemes.

\fakeparagraph{How do our two estimation strategies prioritize steps?}
\label{subsec:correction}
As discussed in \secref{subsec:dde-prioritize}, stepwise and single-shot estimations weaken the optimization of steps around $0$ and $T$ respectively, due to the correction terms and DDIM coefficients becoming larger at different ends.
This phenomenon can be clearly validated in \figref{subfig:correction-coeff}.
Additionally, \figref{subfig:ct_training} shows that the correction terms rapidly converge as training iteration reaches 100, indicating minimal detrimental impact on the training process.

\fakeparagraph{Assigning credits more on the middle part is reasonable.}
We compare the CLIP score of SD15 model under optimizing different parts of the whole denoising trajectory, which is shown in \figref{subfig:analysis}.
The red dashed line denotes the performance of the original SD15 with a score of 0.320. 
Training all denoising steps yields an improvement and the score reaches up to 0.331 (black dashed line). 
By selectively optimizing only 10\% of the denoising steps, the performance fluctuates but does not show clear improvement (blue polylines).
By leveraging the prioritization property inherent in our \algoname method, we specifically optimize only the intermediate subset (steps 200$\sim$700) within the total 1000 denoising steps. This approach yields a further improvement to 0.332 in score. 
These results suggest that even a rough application of the prioritization property, by optimizing only the middle steps, demonstrates significant potential for performance improvement.

\section{Related work}
\label{sec:related}

\fakeparagraph{Reinforcement Learning from Human Feedback (RLHF)}
Using human feedback as a supervision signal in Reinforcement Learning (\cite{MIT98Sutton, ICML18Haarnoja, ICML14Silver}) has been proposed \cite{CAP09Knox, AAAI18Warnell} and evolved to a preference-based paradigm for better performances \cite{ ICML17MacGlashan, NeurIPS17Christiano, NeurIPS18Ibarz}. RLHF has been recently utilized in generative models, particularly large language models \cite{CoRR19Ziegler, CoRR20Stiennon, ICML23Gao}. 
A typical RLHF framework trains a reward model using preference data, and then applies a proximal policy optimization (PPO) \cite{CoRR17Schulman, ICML15Schulman} or other algorithm (\cite{Arxiv24Ethayarajh, NeurIPS24Li, Arxiv24Rosset}) to optimize the policy (\ie, the generative model). 
One key drawback of this approach is the high computational cost.
To address this issue, Direct Preference Optimization (DPO) was proposed \cite{NeurIPS23Rafailov} based on the connection between the optimal policy and the reward model.
This approach has gained significant attention and has been explored in various studies \cite{ICML24Chen, Arxiv24Badrinath, AAAI24Song}. 
However, extending DPO to diffusion models remains an underexplored area.

\fakeparagraph{Alignment training for diffusion model.}
Diffusion models establish a connection between a sample distribution (\eg a Gaussian distribution) and the data distribution \cite{ACM24Yang, ICML24Esser, ICLR24Chen}. 
By learning the inverse process (score matching \cite{UAI09Lyu, NC11Vincent, ICML15Dickstein}), these models can generate images from randomly sampled noise. 
Techniques like latent space conversion \cite{ICLR23Khrulkov, ICLR23Kwon, ICAFG24Haas} and DDIM sampling \cite{ICLR21Songddim} further enhance both the efficiency and effectiveness of the generation process.
Recent studies have explored applying the Direct Preference Optimization (DPO) algorithm to align diffusion models \cite{CVPR24Yang, CVPR24Wallace, ICML24Yang}. 
One of the main challenges in applying DPO to diffusion models is the terminal-only issue of preference labels. 
Existing methods adopt credit assignment perspective \cite{ICML16Wang, ICML16Mnih} and forms two major streams of thought. One stream relies on auxiliary models such as reward models (\cite{ICLR24Black, CoRR23DPOK}) or noisy evaluator (\cite{CoRR24Liang}) which plays counterpart as the noisy classifier in classifier-guided diffusion \cite{NeurIPS21Dhariwal}.
Another stream avoids auxiliary models, using hand-crafting credit assignment schemes such as uniform assignment (\cite{CVPR24Wallace, CVPR24Yang}) or discounted assignment \cite{ICML24Yang} that prioritizes early denoising steps. 

Our method does not rely on auxiliary models therefore it is free from the complexity of extra training.
Furthermore, it manages to avoid the roughly hand-crafted credit assignment scheme and derives a novel scheme by revealing the impact on the terminal denoised distribution of each step.

\section{Conclusion}
\label{sec:conclusion}
In this paper, we introduce Denoised Distribution Estimation (\algoname), a novel direct preference optimization method tailored for diffusion models.
\algoname addresses the challenge of credit assignment across denoising steps, an issue stemming from the terminal-only property of preference labels, by leveraging an enhanced insight into the denoising process. 
\algoname incorporates two estimation strategies that evaluate the impact of each denoising step on the final outcome. 
Our analysis reveals that these two strategies essentially prioritize the optimization of intermediate steps within the denoising trajectory, which is a key distinction from existing methods. 
Extensive experiments demonstrate the effectiveness of our approach compared to previous methods.

\bibliographystyle{ieeenat_fullname}
\bibliography{main}

\maketitlesupplementary

\begin{figure*}[!b]
    \centering
    \includegraphics[width=0.9\linewidth]{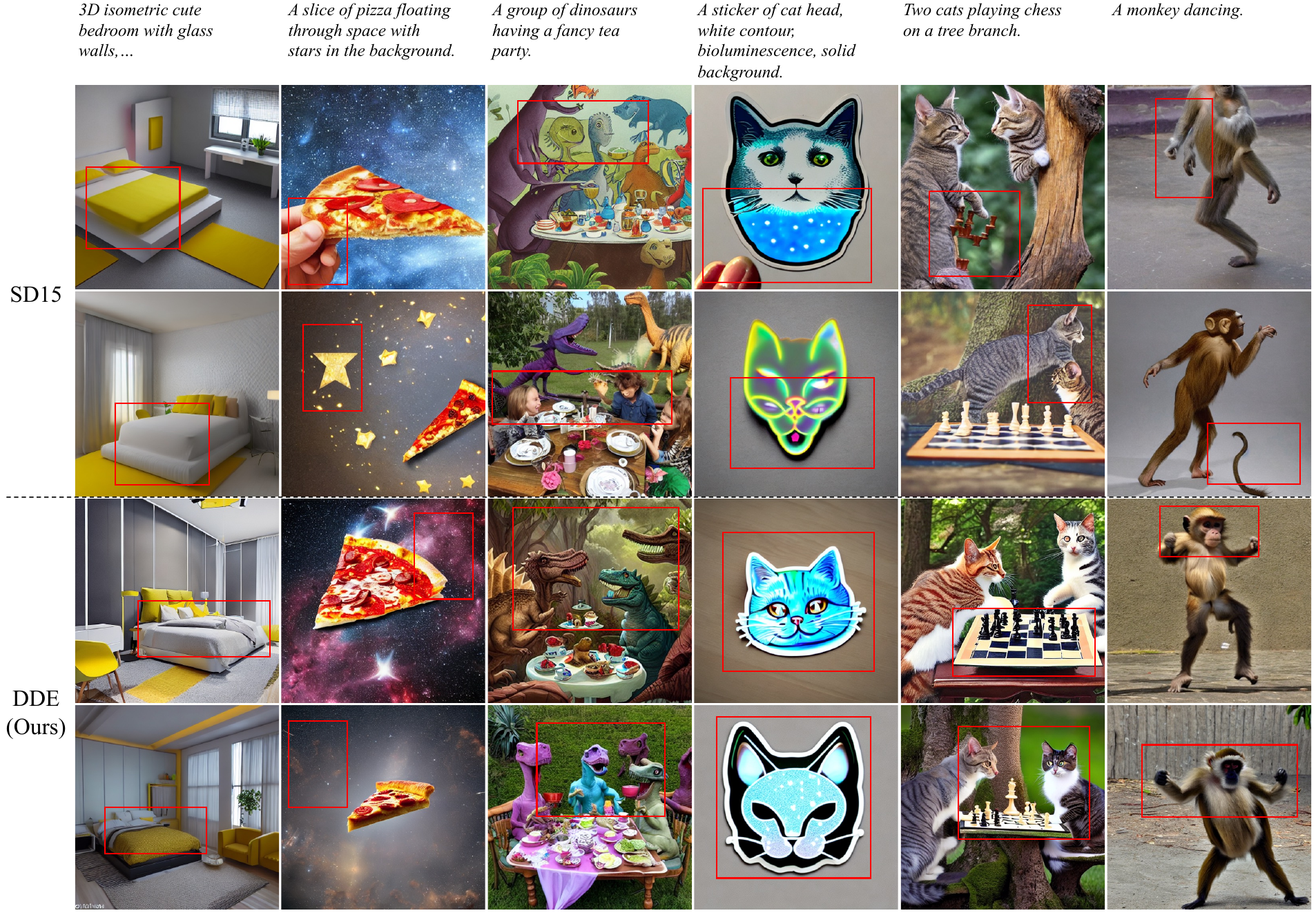}
    \caption{Extended comparative analysis between our \algoname and SD15 model. 
    Our model demonstrates superiority in generating images with enhanced details (\eg beds covered with pillows and quilts, dinosaurs' heads with finer texture), producing more coherent layouts (\eg pizza floating without human hand and cats playing chess at proper position), and avoiding structural collapse (\eg cat-like stickers and the dancing monkey's structure can be retained).}
    \label{fig:cmp-sd15-ext}
\end{figure*}

\section{Derivation}

\subsection{Derivation of the Loss Function Defined in \equref{equ:esa}}
\label{subapp:esa}
By substituting $p_\theta(x_{k}|x_{k+1})$ with $e^{r_k}q(x_k|x_{k+1}, x_0)$ for all $k \in \{t, ..., T-1\}$in \equref{equ:framework}, we obtain:
\begin{equation}
\begin{aligned}
& p_\theta(x_0) = \int_{x_{1:T}} q(x_T)p_\theta(x_{T-1}|x_T)...p_\theta(x_0|x_1) dx_{1:T} \\
&= \exp\{\sum_{k=t}^{T-1} r_k\} \int_{x_{1:T}} q(x_T)\prod_{k=T-1}^tq(x_k|x_{k+1}, x_0) \\
& \prod_{k=t}^1 p_\theta(x_{k-1}|x_k) dx_{1:T}
\end{aligned}
\end{equation}
We now consider the term $q(x_T)\prod_{k=T-1}^tq(x_k|x_{k+1}, x_0)$.
Noting that $q(x_T)$ is independent of $x_0$, we have $q(x_T)=q(x_T|x_0)$.
Applying Bayes' theorem, it follows that:
\begin{equation}
\begin{aligned}
& \int q(x_T|x_0)q(x_{T-1}|x_T, x_0)... dx_{t:T-1} dx_T \\
&= \int q(x_T, x_{T-1}|x_0)dx_T ... dx_{t:T-1}\\
&= \int q(x_{T-1}|x_0) q(x_{T-2}|x_{T-1},x_0)... dx_{t:T-2} dx_{T-1} \\
&=\int q(x_{T-2}|x_0)... dx_{t:T-2} \\
&=\int q(x_t|x_0) dx_t \\
\end{aligned}
\end{equation}
Thus, we obtain:
\begin{equation}
\begin{aligned}
p_\theta(x_0) 
&\approx \exp\{\sum_{k=t}^{T-1} r_k\} \int_{x_{1:t}} q(x_t|x_0) \prod_{k=t}^1 p_\theta(x_{k-1}|x_k) dx_{1:t} \\
\end{aligned}
\end{equation}

\begin{figure*}[b]
    \centering
    \includegraphics[width=0.9\linewidth]{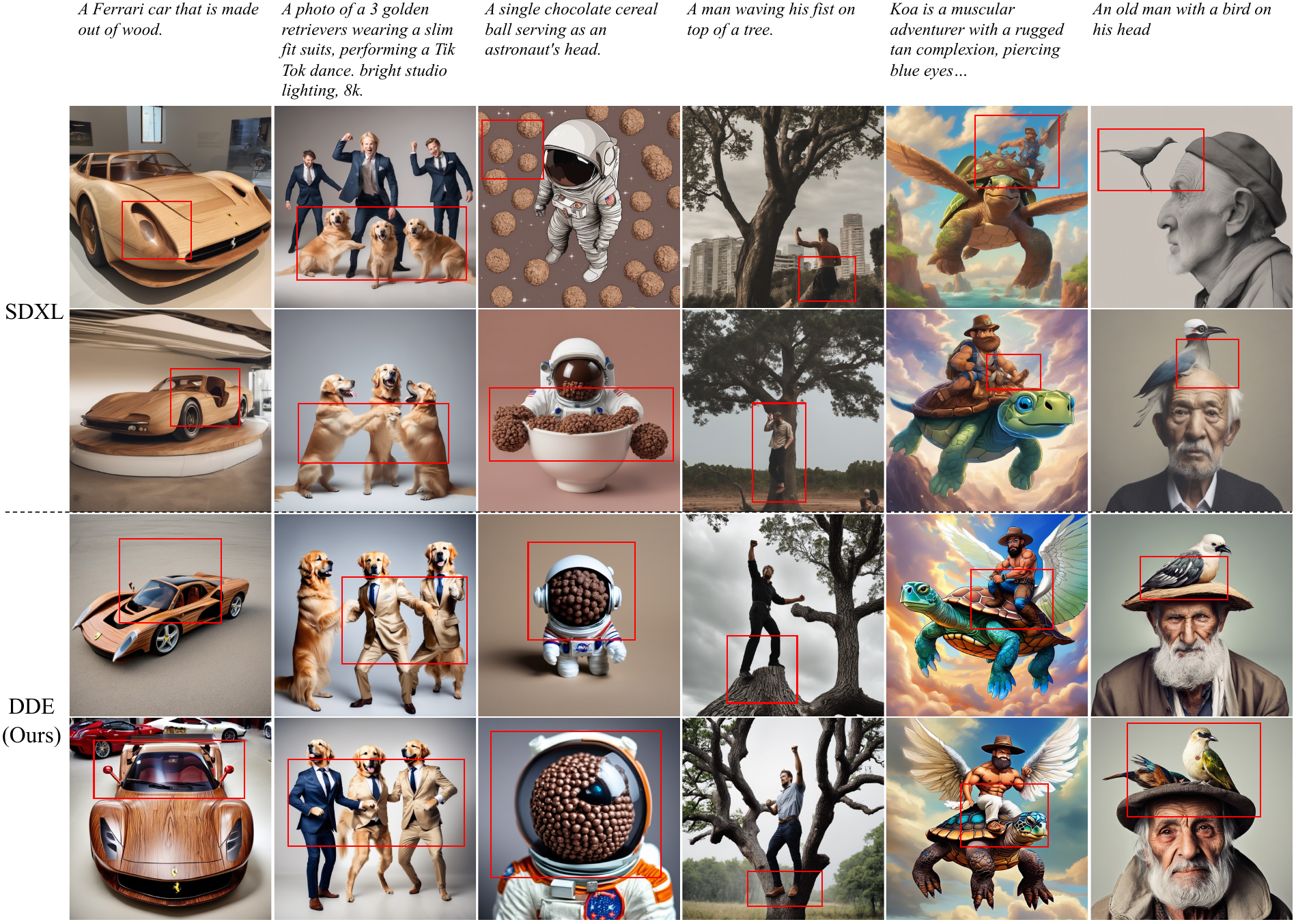}
    \caption{Extended comparative analysis between our \algoname and SDXL model. The results indicate our \algoname excels in producing more intricately detailed car models.
    Additionally, \algoname exhibits a heightened ability to follow prompt instructions (\eg the dogs wearing slim suits, the astronaut's head shaping like a cereal ball, and the man waving fists on a tree).
    Furthermore, our model consistently maintains accurate structural integrity in interfaces, such as those between cartoon figure and the turtle, as well as between birds and human heads.}
    \label{fig:cmp-sdxl-ext}
\end{figure*}

\subsection{Derivation of \equref{equ:dpo-loss-approx}}
\label{subapp:loss}
By replacing $p_\theta(x_0)$ and $p_{ref}(x_0)$ in the logarithmic term of \equref{equ:dpo-naive} with \equref{equ:all}, we obtain:
\begin{equation}
\begin{aligned}
& \log \frac{p_\theta(x^w_0)}{p_{ref}(x^w_0)} \\
&= \sum_{k=t}^T (r^w_{\theta, k} - r^w_{ref, k}) + \log \frac{\mathbb{E}_{x^w_t \sim q(x^w_t|x_0)}[p_\theta(x^w_0|x_t)]}{\mathbb{E}_{x^w_t \sim q(x^w_t|x_0)}[p_{ref}(x^w_0|x_t)]}
\end{aligned}
\end{equation}
To avoid the high computation cost of the integral calculation related to the expectation, we employ the Monte Carlo method, using a single point $x_t\sim q(x_t|x_0)$ for estimation. Applying this estimation as model input to both the target and reference models, we derive:
\begin{equation}
\begin{aligned}
& \log \frac{\mathbb{E}_{x^w_t \sim q(x^w_t|x_0)}[p_\theta(x^w_0|x_t)]}{\mathbb{E}_{x^w_t \sim q(x^w_t|x_0)}[p_{ref}(x^w_0|x_t)]} \\
& = -||x^w_0 - \hat{\mu}_{\theta, t'=0}(x^w_t, t)||_2^2 + ||x^w_0 - \hat{\mu}_{ref, t'=0}(x^w_t, t)||_2^2 
\end{aligned}
\end{equation}

The derivations of the terms $\frac{p_\theta(x_0^w)}{p_{ref}(x_0^w)}$ and $\frac{p_\theta(x_0^l)}{p_{ref}(x_0^l)}$ follow the same procedure.
Consequently, the total loss function is given by:
\begin{equation}
\begin{aligned}
& \mathcal{L}_{\text{\algoname}}  \\
&= \mathbb{E}_{x^w_0, x^l_0, x^w_t \sim q(x^w_t|x^w_0), x^l_t \sim q(x^l_t|x^l_0)}[-\log \sigma(\beta(  \\
& -||x_0^w - \hat{\mu}_{\theta, t'=0}(x_t^w, t)||_2^2 + ||x_0^w - \hat{\mu}_{ref, t'=0}(x_t^w, t)||_2^2  \\
& + ||x_0^l- \hat{\mu}_{\theta, t'=0}(x_t^l, t)||_2^2 - ||x_0^l - \hat{\mu}_{ref, t'=0}(x_t^l, t)||_2^2  \\
& +\sum_{k=t}^T (r_{\theta,k}^w - r_{ref, k}^w - r_{\theta, k}^l + r_{ref, k}^l)
))]
\end{aligned}
\end{equation}

\section{Extended Experiments}
\label{app:exp}

\subsection{Implementation Details}
\label{subapp:detail}
We employ a constant learning rate with a warm-up schedule, finalizing at $2.05 \times 10^{-5}$.
The hyper-parameters $\beta$ and $\mu$ are set to $5000$ and $0.1$ respectively.
To optimize computational efficiency, both gradient accumulation and gradient checkpointing techniques are utilized.
The validation set of Pick-a-pic dataset contains over 300 prompts. For each prompt, we generate eight images and subsequently evaluate the scores using various models.


\subsection{Extended Qualitative Evaluation}
\label{subapp:quality}

In this subsection, we present additional generated case comparisons to substantiate the superior quality of our method.

A comparison of the SD15 model can be found in \figref{fig:cmp-sd15-ext}.
Our model demonstrates superior proficiency in generating intricate content. For instance, the beds produced by our model display quilts and pillows with finer folds.
Moreover, the images generated by our method exhibit increased coherence, as the pizza appears to float without the presence of human hands and the background stars exhibit a more natural look.
The dinosaurs' heads our model generate possess finer structures and textures compared to the base model.
The cat stickers generated by our \algoname maintain the structural integrity of the cat head.
Additionally, our model excels in generating images with a more appropriate layout, as evidenced by the two cats playing chess with a clearer chessboard and more natural positions than those generated by the base model.
The dancing monkeys created by our method better preserve body structures than those produced by the base model.

The comparative analysis of the SDXL model is depicted in  \figref{fig:cmp-sdxl-ext}.
For cars made out of woods, our model exhibits superior detail generation.
It also demonstrates enhanced comprehension of prompts, accurately depicting dogs \textit{wearing} suits, and astronauts with heads composed of cereal balls, as specified.
Our generated human figures exhibit postures and positions that better adhere to the prompt requirements compared to those produced by the SDXL model.
Additionally, our model effectively generates cartoon figures, avoiding structure collapse and providing a clearer interface between the man's head and the bird.

\end{document}